\documentclass[letterpaper, 10 pt, conference]{ieeeconf}  

\IEEEoverridecommandlockouts                              

\usepackage{amsmath} 
\usepackage{amssymb}  
\usepackage[ruled,linesnumbered]{algorithm2e}
\usepackage{algorithmic}
\usepackage{mathrsfs}
\usepackage{url}
\usepackage{color}
\usepackage{xcolor}
\usepackage{cite}
\usepackage{multirow}
\usepackage{xspace}
\usepackage{graphicx} 
\usepackage{subfigure}
\usepackage{makecell}
\usepackage{threeparttable}

\newcommand{\core}{\emph{MAROAM}\xspace}
\newcommand{\pfilter}{\emph{pFilter}\xspace}

\makeatletter
\let\NAT@parse\undefined
\makeatother
\usepackage[colorlinks,
            urlcolor=blue,
            linkcolor=blue,       
            anchorcolor=blue,  
            citecolor=green        
            ]{hyperref}

\title{\LARGE \bf
\core: Map-based Radar SLAM through Two-step Feature Selection
}

\author{Dequan Wang, Yifan Duan, Xiaoran Fan, Chengzhen Meng, Jianmin Ji and Yanyong Zhang
\thanks{School of Computer Science and Technology, University of Science and Technology of China, Hefei, 230026, China
{\tt\small \{wdq15588, dyf0202, czmeng\}@mail.ustc.edu.cn, ox5bc@winlab.rutgers.edu, \{jianmin, yanyongz\}@ustc.edu.cn}.}%
}%

\begin{document}

\maketitle
\thispagestyle{empty}
\pagestyle{empty}

\begin{abstract}

In this letter, we propose  \core, a millimeter wave radar-based SLAM framework, which employs a two-step feature selection process to build the global consistent map. Specifically, we first extract feature points from raw data based on their local \emph{geometric} properties to filter out those points that violate the principle of millimeter-wave radar imaging. Then, we further employ another round of \emph{probabilistic} feature selection by examining how often and how recent the feature point has been detected in the proceeding frames. With such a two-step feature selection, we establish a global consistent map for accurate and robust pose estimation as well as other downstream tasks. At last, we perform loop closure and graph optimization in the back-end, further reducing the accumulated drift error.

We evaluate the performance of \core on the three datasets: the Oxford Radar RobotCar Dataset, the MulRan Dataset and the Boreas Dataset. We consider a variety of experimental settings with different scenery, weather, and road conditions. The experimental results show that  the accuracy of \core is 7.95\%, 37.0\% and 8.9\% higher than the currently best-performing  algorithms on these three datasets, respectively. The ablation results also show that our map-based odometry performs 28.6\% better than the commonly used scan-to-frames method. Finally, as devoted contributors to the open-source community, we will open source the algorithm after the paper is accepted.

\end{abstract}

\section{INTRODUCTION} \label{sec:intro}


In recent years, millimeter wave radar (radar, in short) has received an increasing amount of attention in autonomous systems. Compared to vision and LiDAR, radar offers longer detection ranges and more robust performance under adverse lighting and weather conditions, lending it towards a viable sensing option for applications like self driving and robotics\cite{cen2020a}. However, in order for radar to grow from a supplementary sensing approach to one of the primary ones, their sensing capability, especially when used alone, needs to be carefully investigated and strengthened.  Towards this goal, 
we study how to utilize the radar alone for the Simultaneous Localization and Mapping (SLAM) task in this work.

The main weakness for millimeter wave radar images lies in the fact that they suffer from the multipath effect, sidelobes, clutter, etc.\cite{cen2020a}, which may lead to false alarms.
In order to overcome the effect of false alarms,  the recent radar SLAM systems mainly focus on extracting robust features from radar images. For example, Cen et al.  proposed to extract features using traditional signal processing techniques\cite{8460687,8793990}, while Barnes et al. proposed end-to-end learning-based feature extraction methods\cite{9196835,maskbymoving}. However, 
even with these efforts, feature extraction for radar systems remains an active research topic,  which aims at minimizing the cumulative errors in the feature matching task. 
\begin{figure}[t]
	\centering
	\includegraphics[width = \linewidth]{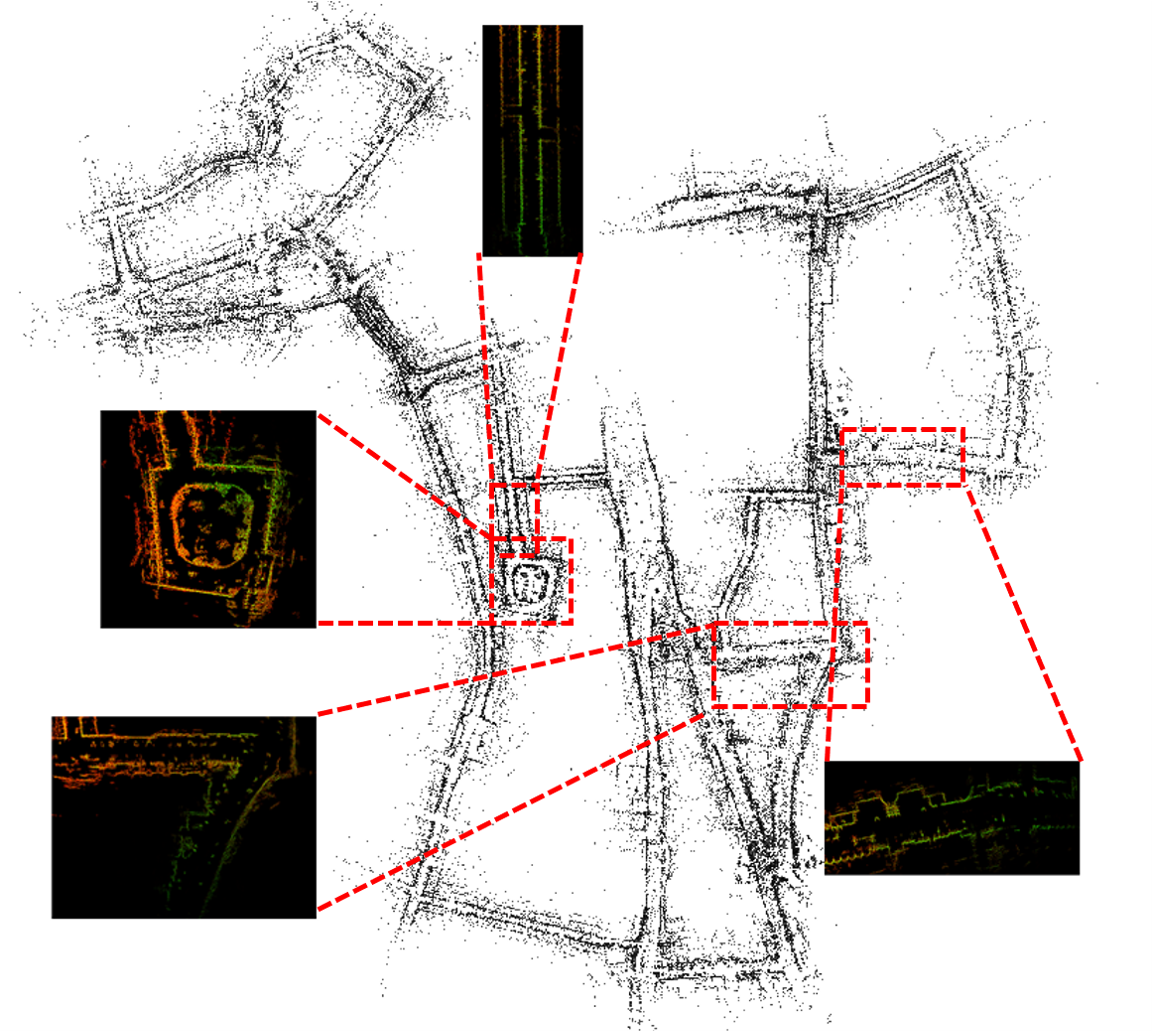}
	\caption{An example radar point cloud map with global consistency, which is built by \core with the Oxford Radar RobotCar Dataset. For clarity, we also zoom in several locations and show the detailed views. In this example, we use the ''17-13-26'' sequence with a total length of $10km$.}
	\label{pictures:1}
\end{figure}

Next, let us take a close look at the radar odometry. When the scan-to-map registration strategy is engaged, the number of features grows rapidly with the map, a large portion of which are of low quality and should not be selected. The quality of the map seriously affects the performance of map based-SLAM\cite{7747236}. RadarSLAM\cite{RadarRobotCarDatasetICRA2020} uses descriptor-based pairwise matching of features. However, since image noise and different radar cross-sectional (RCS) generated by different viewing angles\cite{borkar2010radar} will directly affect the representation of descriptors, so it has high requirements for the extraction process of features and descriptors. 

Since pairwise matching of descriptors is not required, the Iterative Closest Point (ICP)\cite{Besl1992AMF} methods achieve state-of-the-art accuracy in motion estimation, such as CFEAR RadarOdometry\cite{adolfsson2021cfear} and NDT\cite{9561413}, but the submaps they use are only the superposition of multiple frames, and there is no mapping process. The disadvantage of ICP-based point cloud registration algorithms is that they are sensitive to initial values and tend to converge to local optimum\cite{review2021}. Since radar point cloud has more noise, the 
ICP-based algorithm is more likely to fall into a local optimum, resulting in a larger motion estimation error. We have demonstrated this experimentally at \ref{Ablation}. Considering the importance of map quality, our paper looks at how feature selection can be used to optimize the process of map building and scan-to-map matching in ICP-based radar SLAM, resulting in more accurate maps and more robust state estimates.

In order to address these limitations, we propose a Map-based Radar Odometry and Mapping system, \core, which is an accurate and robust map-based radar SLAM framework based on LOAM\cite{LOAM1}. In order to improve map quality, we use a two-step geometry-probability feature selection strategy in the registration and map update stage. In specificity, we first extract surface features by calculating the Local Linearity and the Local Aggregation with a \emph{geometry-based} filter. We then involve a point-to-line ICP method, using the surface features to solve the relative pose transformation. Subsequently, we use a \emph{probability-based} filter to dynamically filter out the features that are not frequently selected in the ICP matching process. Using the more frequent features, our algorithm realizes the scan-to-map matching strategy of ICP-based radar SLAM for the first time. Additionally, We use the Scan Context\cite{8593953} for loop closure and modify it to fit the radar data format.


We evaluate the performance of \core on the three datasets: the Oxford Radar RobotCar Dataset\cite{RadarRobotCarDatasetICRA2020}, the MulRan Dataset\cite{9197298} and the Boreas Dataset\cite{burnett_boreas22}. We consider a variety of experimental settings with different scenery, weather, and road conditions. The experimental results show that  the accuracy of \core is 7.95\%, 37.0\% and 8.9\% higher than the currently best-performing  algorithms on these three datasets, respectively. The ablation results also show that our map-based odometry performs 28.6\% better than the commonly used scan-to-frames method. 

In summary, our main contributions are as follows:
\begin{itemize}
    \item We propose \core, an map-based radar SLAM framework based on LOAM. Using a geometry-probability two-step feature selection, our algorithm realizes the scan-to-map matching strategy of ICP-based radar SALM for the first time.
    \item We evaluate our \core on three datasets which contain a variety of scenarios, weathers, and road conditions. Experiments show that our algorithm outperforms SOTA algorithm on all three datasets.
\end{itemize}


\section{Related Works}

\subsection{Radar SLAM}

We divide the existing radar odometry/SLAM algorithms into four categories according to the method of estimating the relative pose:

\subsubsection{Direct methods}Checchin et al. \cite{Checchin} first time use the Fourier-Mellin Transform (FMT) to register radar images in a sequence for motion estimation. PhaRao\cite{9197231} apply FMT to Cartesian and log polar radar images to estimate rotation and translation to decouple rotation and translation.

\subsubsection{ICP-based methods}Adolfsson et al.\cite{adolfsson2021cfear} process the landmarks extracted by CFAR through k-strongest filtering in Cartesian coordinates, and optimizes the point-to-line optimize metric to estimate the relative pose. Kung et al.\cite{KungNDT} propose a RO with probabilistic submap building, and an NDT-based radar scan matching. Both of them use scan-to-submap matching, but the submaps they use are only the superposition of multiple frames, and there is no mapping process.

\subsubsection{Descriptor-based methods}
Considering the unnecessary influence of radar echo, Cen et al.\cite{8460687} propose an algorithm to extract landmarks, and performe scan matching by greedily adding features correspondence based on unary descriptor and pairwise compatibility score. After that, another feature extraction algorithm that only use one parameter is proposed by Cen et al.\cite{8793990}, and graph matching is used for scan matching. Hone et al.\cite{Hong2021RadarSA,RadarSLAMv2} propose a full radar-based SLAM pipeline, RadarSLAM, composed of pose tracking, local mapping, loop closure detection and pose graph optimization. RadarSLAM uses visual features for scan matching and M2DP\cite{7759060} descriptor based on point cloud for loop detection.

\subsubsection{Learning-based methods}Barnes et al.\cite{maskbymoving} use deep neural network to learn an embedding space that is basically free of artifacts and interference, which is used to perform effective correlation matching between continuous radar scans, and achieve high accuracy without considering spatial cross-validation(CSV). A self supervised learning framework is proposed by Barnes et al.\cite{9196835} to detect the robust key features of range estimation and metric positioning in radar. Burnett et al.\cite{HERO} uses unsupervised method to extract features, which increases generalization and performs well on both Oxford and Boreas.

\subsection{LiDAR SLAM}

LOAM\cite{LOAM1} is the pioneering work of 3D LiDAR SLAM, which proposes a basic framework. LOAM extracts edge features and plane features from the original point cloud, and designs related loss functions for each type of feature. The matching consists of a fast frame-to-frame match and a slow frame-to-graph match. F-LOAM\cite{wang2021} follows LOAM and abandons the scan-to-scan match and replaces it by only scan-to-map with high frequency. Duan et al.\cite{pfilter} propose a feature filter, \pfilter, by properly measuring each feature point’s p-Index and only keeping those with high index values, and improved both the efficiency and accuracy of the registration process. Our work is inspired by the aforementioned LiDAR SLAM.



\begin{figure}[t]
	\centering
	\includegraphics[width = \linewidth]{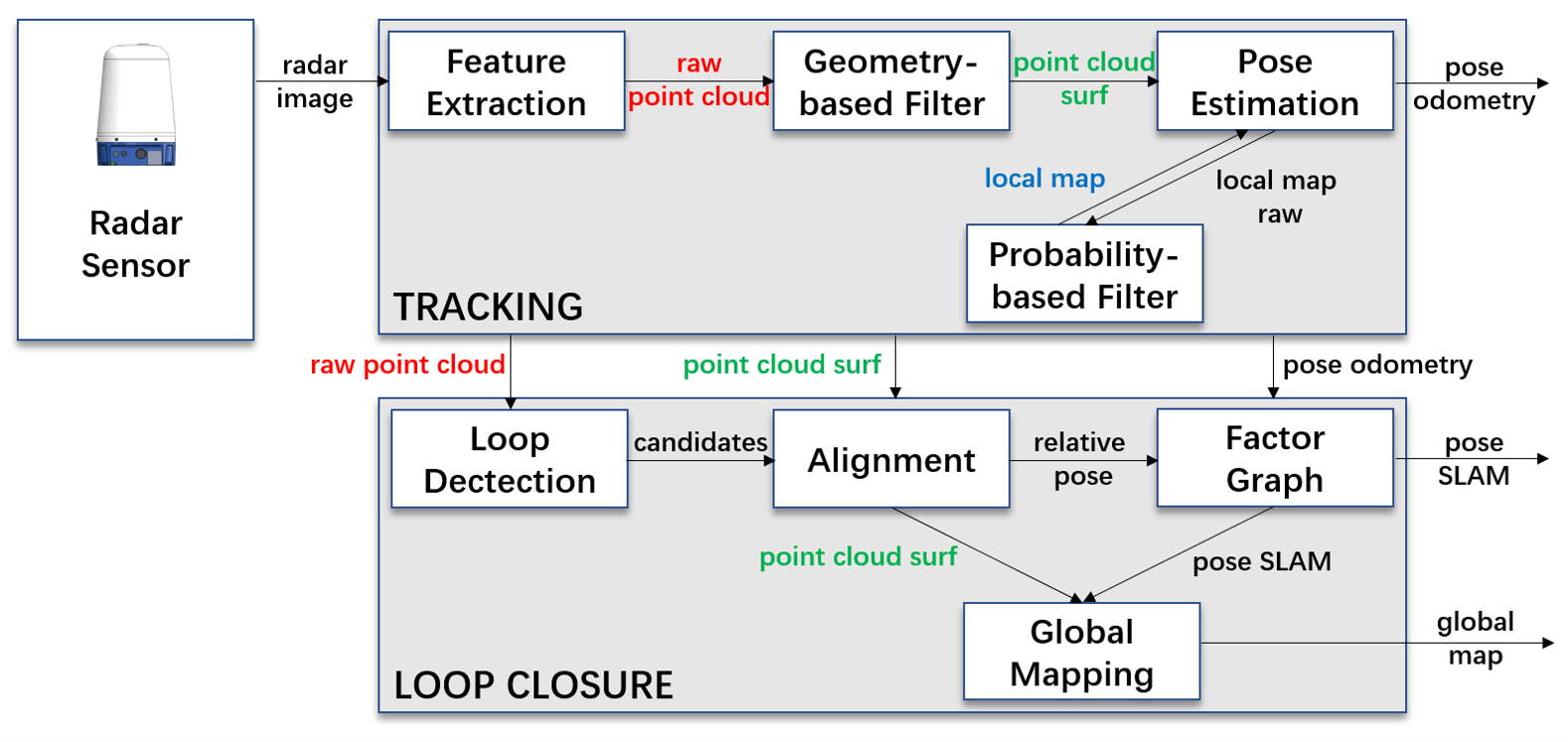}
	\caption{Pipeline of the \core. The \textbf{raw point cloud} represents the point cloud directly extracted by the feature extractor, \textbf{point cloud surf} represents the surface features filtered by \emph{geometry-based} filter, \textbf{raw local map} is the union of the local map and the current \textbf{point cloud surf}, \textbf{local map} is generated from \textbf{raw local map} filtered by \emph{probability-based} filter.
	}
	\label{fig:pipline}
\end{figure}
\section{Method}
\begin{figure}[t]
	\centering
	\subfigure[raw point cloud] {\includegraphics[width = 0.48\linewidth]{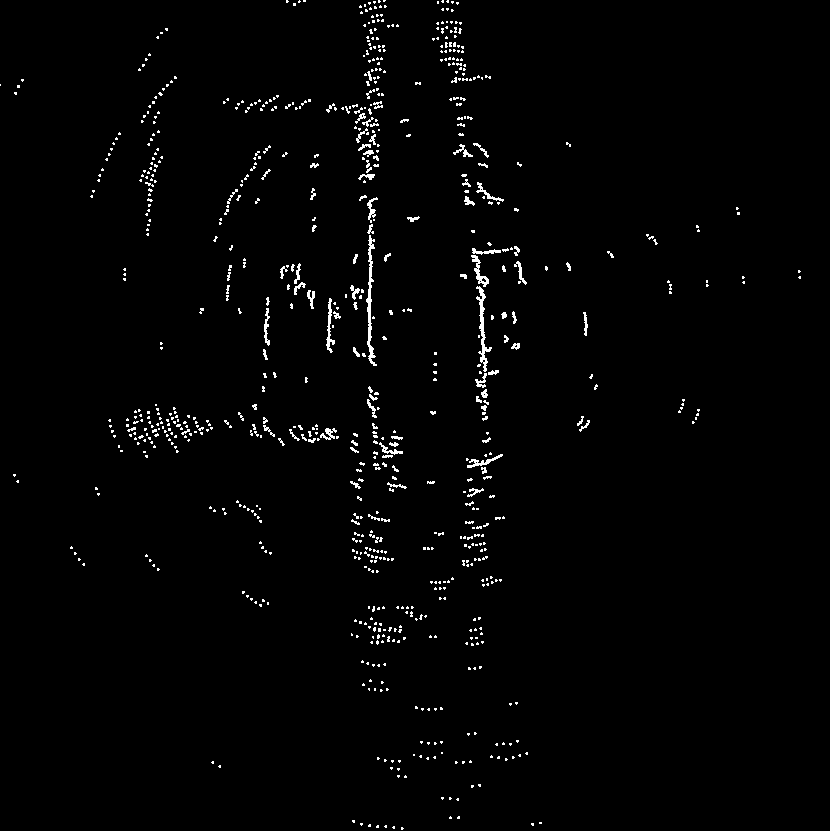}}
	\subfigure[point cloud surf] {\includegraphics[width = 0.48\linewidth]{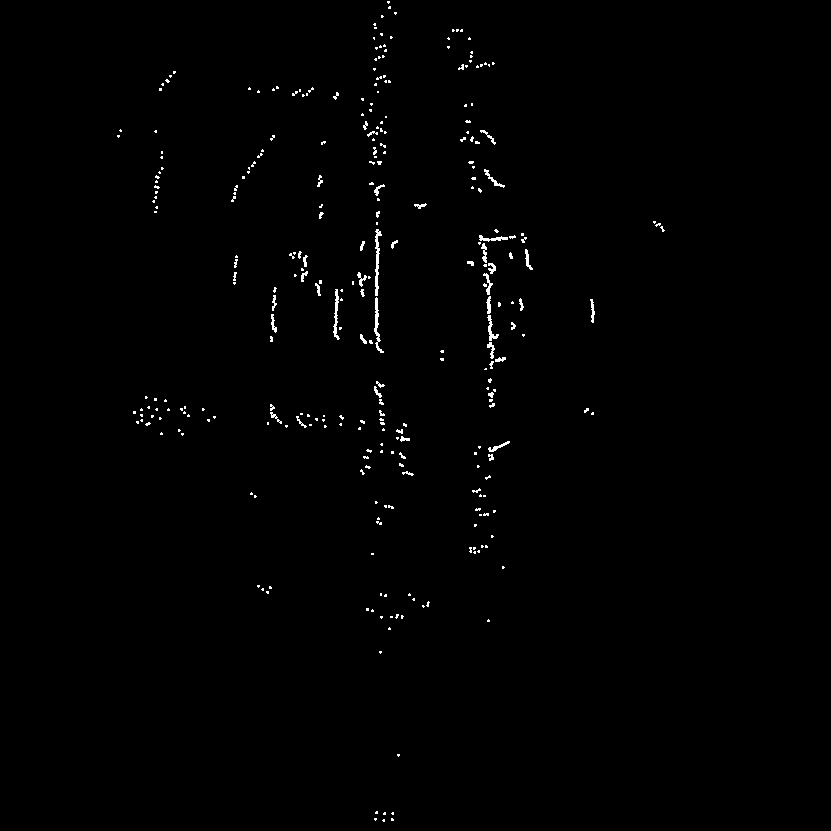}}
	\subfigure[Local map before \newline probability-based filter] {\includegraphics[width = 0.48\linewidth]{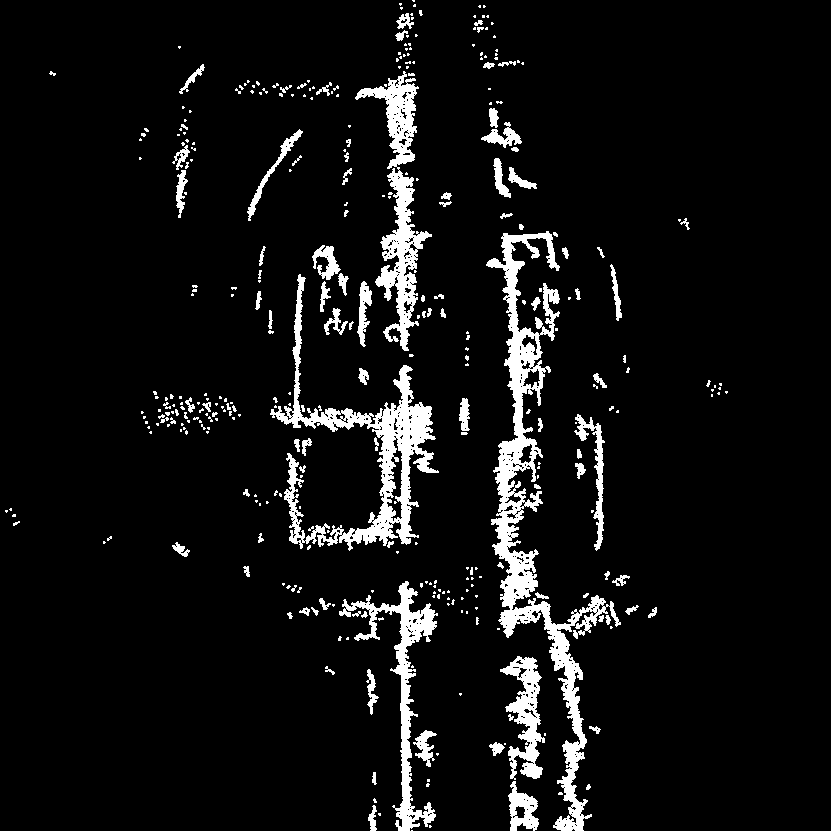}}
	\subfigure[Local map after \newline probability-based filter] {\includegraphics[width = 0.48\linewidth]{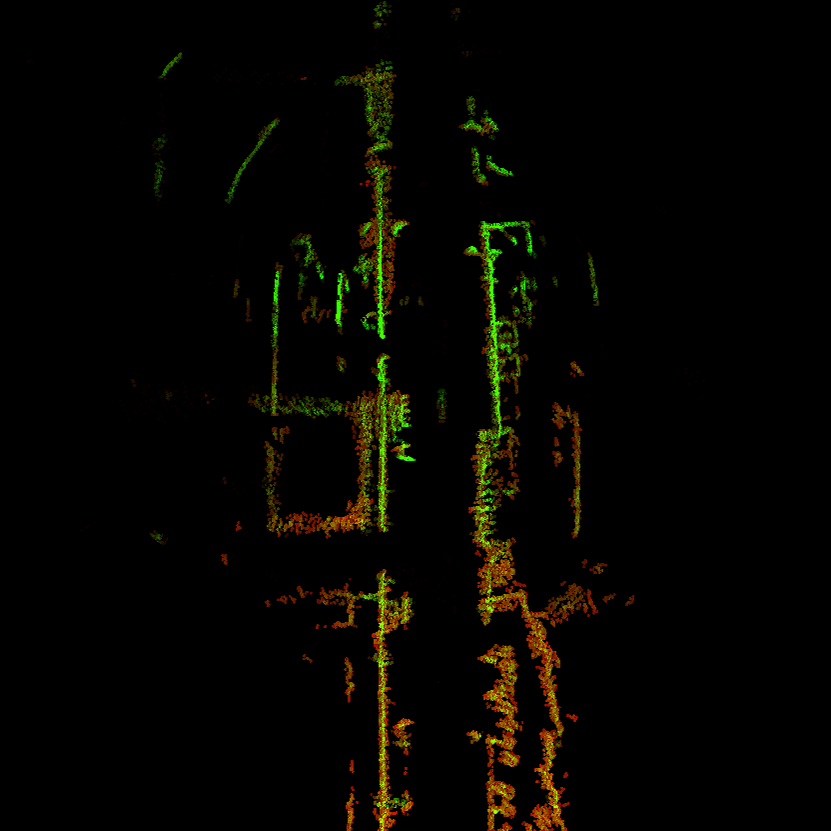}}
	\caption{The effect of \emph{geometry-based} filter and \emph{probability-based} filter for a radar scan and a local feature map. The raw point cloud (a) contain a large number of wrong feature points such as ghosts and speckles; (b) shows the point cloud after \emph{geometry-based} filter; (c) shows the local surface map by stacking all surface feature points; (d) shows local surface map after  \emph{probability-based} filter. Each feature point in the maps is colored by its hit probability value: the red color indicates feature points with low hit probability values, i.e., transient feature points, while the green color indicates feature points with high hit probability values, i.e.,  frequent feature points. The resulting maps in (d)  contain fewer feature points that should be included for registration.}
	\label{fig:show}
\end{figure}

\subsection{System Overview}

An overview of \core is shown in Fig.~\ref{fig:pipline}. The system receives input from a 2D radar image and outputs a 3-DOF pose estimation and a global consistent map. The SLAM system consists of two modules: tracking and loop closure. The tracking module is used for feature extraction, feature filtering and real-time motion estimation, and the loop closure module is used for loop detection and graph optimization.

\subsection{Geometry-based Feature Filter}
Cen2019\cite{8793990} is used as feature detector in \core. Feature detector firstly identifies continuous regions with high intensity and low gradient for each azimuth of a radar image. Then, the midpoints of each continuous region are extracted as features. 
The point cloud extracted from the \textbf{k}th scan is denoted as $\mathcal{P}_k$. 

The \emph{geometry-based} filter is used to extract surface features from raw point cloud, marked as $\mathcal{S}_k$. Surface features are defined as points with high aggregation and high local linearity. Firstly, we use the principal components analysis (PCA) to calculate the the local linearity. For each feature $\textbf{p}_i\in \mathcal{P}_k$, we find the nearest $\textbf{m}$ points to $\textbf{p}_i$ from the point cloud $\mathcal{P}_k$ to form a corresponding set, marked $\mathcal{N}_i$. We store the point cloud $\mathcal{P}_k$ in a KD-tree to accelerate nearest neighbor search. Let $\Bar{\textbf{p}}_i$ be the centroid of $\mathcal{N}_i$ , and $\mathbf{C}_i$ be the covraiance matrix of $\mathcal{N}_i$ which is calculated as
\begin{equation}
\mathbf{C}_i=\frac{1}{|\mathcal{N}_i|}\sum_{\textbf{p}_j\in \mathcal{N}_i}(\textbf{p}_j -\Bar{\textbf{p}}_i)((\textbf{p}_j -\Bar{\textbf{p}}_i)^T,
\end{equation}
Then We use PCA to calculate the eigenvalues of matrix $\mathbf{C}_i$. The largest two eigenvalues are recorded as $\lambda_1$, $\lambda_2$($\lambda_1>\lambda_2$). We define local linearity $\theta$ as
\begin{equation}
\theta = \frac{\lambda_1 - \lambda_2}{\lambda_1}.
\end{equation}
Lastly, a lower bound $\theta_{min}$ is set to select features with high local linearity. In order to extract features with high aggregation degree, we compute the radius(marked as $\textbf{r}$) of $\mathcal{N}_i$, which is the maximum distance between the feature and the point in $\mathcal{N}_i$. If a feature's local linearity $\theta > \theta_{min}$ and radius $\textbf{r}$ is less than the maximum distance $\textbf{d}_{max}$, we consider it as a surface feature point and keep it. We set $\textbf{m}=10$, $\textbf{d}_{max}=2$, $\theta_{min}=0.9$.

\subsection{Pose Tracking}
\subsubsection{Motion Compensation}
Since the radar feature points in a scan are received at different times, we need to align all feature points to the beginning time of the scan. The process is called Motion Compensation. As an approximation, the radar motion is modeled with constant angular and linear velocities between two scans. This allows us to use linear interpolation to calculate the positions of features at the beginning of the scan. 



\subsubsection{Motion Estimation}

The pose estimation aligns the current surface feature point cloud $\mathcal{S}_k$ with the global feature map $\mathcal{M}_{k-1}$. We solve the pose (marked as $T_k$) of the \textbf{k}th scan by optimizing the loss function of the distance between surface feature point cloud $\mathcal{S}_k$ and the global feature map $\mathcal{M}_{k-1}$, and the distance is calculated by point-to-line metric:
\begin{align}
T_k &= \arg \min_T f(\mathcal{M}_{k-1},\ \mathcal{S}_k,\ T)\notag\\
 &=  \arg \min_T\sum_{\textbf{p}_i^S\in\mathcal{S}_k} ||((T\textbf{p}_i^S-\textbf{p}_i^M)\times \vec{n}_i||^2,
\end{align}
where $\textbf{p}_i^S$ is a surface feature in $\mathcal{S}_k$, $\textbf{p}_i^M$ is the geometric center of the corresponding points set in global feature map $\mathcal{M}_{k-1}$, and $\vec{n}_i$ is the unit vector of the norm of the corresponding points set. 





\subsection{Probability-based Feature Filter}


The feature points selected by geometry-based feature selection still contain some noise caused by multi-path effect and power saturation. These noise has little effect when using scan-to-scan or scan-to-frames matching strategy. However, when \core need to build a local map and use scan-to-map matching strategy, wrong features will be accumulated in the local map, which will significantly reduce the accuracy of motion estimation.

Inspired by \pfilter\cite{pfilter}, we design a probability-based feature filter to further extract more stable feature points. \pfilter propose a new metric ''p-Index'' to evaluate feature points, which indicates whether a point is frequent or transient. With the same thought, we define a simple metric as the frequency of a feature point being hit.


For a feature point $p_{k_0}^M$ that appears for the first time in the $k_0$th scan map, we define $\textbf{I}(\textbf{p}_{k_0}^M, k)$ to indicate whether $\textbf{p}_{k_0}^M$ is \textit{hit} in the $k$th ($k>k_0$) scan:
\begin{equation}
    \textbf{I}(\textbf{p}_{k_0}^M, k) = \left\{\begin{array}{ll}
    1 &  \exists\,\textbf{p}_k^S\in \mathcal{S}_k, s.t.\  \textbf{p}_{k_0}^M\in \mathcal{N}_k(\textbf{p}_k^S),\\
    0 & \text{otherwise},
    \end{array}
    \right.
\end{equation}
where $\mathcal{S}_k$ denotes point cloud of extracted surface feature points from the $k$th images of radar, and $\mathcal{N}_k(\textbf{p}_k^S)$ denotes the set of corresponding feature points in $\mathcal{M}_k$ for $\textbf{p}_k^S$ in $S_k$. Intuitively, $\textbf{I}(\textbf{p}_{k_0}^M, k) = 1$ indicates that there exits an extracted feature point in the $k$th scan that can be matched with the point $\textbf{p}_{k_0}^M$, and it is called to be \textit{hit}.

For each feature point $\textbf{p}_{k_0}^M$ in the feature map $\mathcal{M}_k$, we define $R(\textbf{p}_{k_0}^M, k)$ and $H(\textbf{p}_{k_0}^M, k)$ to describe the attributes of feature points. $R(\textbf{p}_{k_0}^M, k)$ represents the matching rounds of the feature point:
\begin{equation}
R(\textbf{p}_{k_0}^M, k)=k-k_0+1.
\end{equation}
And $H(\textbf{p}_{k_0}^M, k)$ represents the total number of hits of the feature in all historical matches:
\begin{equation}
H(\textbf{p}_{k_0}^M, k)=\sum_{k'=k_0+1}^k \textbf{I}(\textbf{p}_{k_0}^M, k').
\end{equation}
The above formulas can be expressed in recursive form:
\begin{equation}
\begin{array}{ll}
    R(\textbf{p}_{k_0}^M, k+1)&=R(\textbf{p}_{k_0}^M, k)+1,\\
    H(\textbf{p}_{k_0}^M, k+1)&=H(\textbf{p}_{k_0}^M, k)+\textbf{I}(\textbf{p}_{k_0}^M, k+1).
    \end{array}
\end{equation}
At last, we define the hit probability $P(\textbf{p}_{k_0}^M, k)$ for each feature $\textbf{p}_{k_0}^M$ as the hit number divide the round number:
\begin{equation}
P(\textbf{p}_{k_0}^M, k)=H(\textbf{p}_{k_0}^M, k)/R(\textbf{p}_{k_0}^M, k).
\end{equation}

For feature points with matching map feature points in the neighborhood, we estimate an initial value of $R(\textbf{p}_{k_0}^M, k_0)$ and $H(\textbf{p}_{k_0}^M, k_0)$ for each feature point $\textbf{p}_{k_0}^M$, using the average value of $R$ and $H$ for five feature points that correspond to $\textbf{p}_k^M$ in the match. 

We define three parameters $\theta_p$, $R_{min}$, $H_{max}$ for feature selection:

\begin{itemize}
    \item If $\textbf{p}_{k_0}^M$ has just been detected recently, i.e., $R(\textbf{p}_{k_0}^M, k) < R_{min}$, then we maintain it in the local map for a while to see whether it would be stable;
    \item If a feature $\textbf{p}_{k_0}^M$ is detected many times, i.e., $H(\textbf{p}_{k_0}^M, k) > H_{max}$, we still consider it a stable feature even if it is not detected for many scans afterward;
    \item At last, $\theta_p$ represents as the threshold to identify stable feature points. For a feature $\textbf{p}_{k_0}^M$, If $H(\textbf{p}_{k_0}^M, k_0) > \theta_p$, it will be maintained in the local feature map permanently. Otherwise, it will be removed from local feature map.
\end{itemize}

The algorithm will select features in two stages: feature matching and map update. The process of probability-based feature selection is summarized in Algorithm \ref{alg1}.

\begin{algorithm}[t]
\caption{Probability-based Feature Filter}\label{alg1}
\KwData{$\mathcal{M}_k,\ \mathcal{S}_k,\ \textbf{T}_k$}
\KwResult{$\mathcal{M}_{k+1}$}
\For{$\textbf{p}_k^S \emph{\textbf{in}} \mathcal{S}_k$}{  
    $\textbf{p}_k^S \leftarrow \textbf{T}_k\textbf{p}_k$\;
    $\mathcal{N}_k(\textbf{p}_k) \leftarrow Find\_correspondence(\textbf{p}_k^S, \mathcal{M}_k,5$)\;
    \For{$\textbf{p}_{k_0}^M \textbf{in} \mathcal{N}_k(\textbf{p}_k^S)$}{
        $H(\textbf{p}_{k_0}^M,k)\leftarrow H(\textbf{p}_{k_0}^M,k) + 1$\;
    }
    $H(\textbf{p}_k^S,k) \leftarrow Average(H(\textbf{p}_{k_0}^M,k)$ 
    \textbf{for} $\textbf{p}_{k_0}^M \textbf{in} \mathcal{N}_k(\textbf{p}_{k}^S))$\;
    $R(\textbf{p}_k^S,k) \leftarrow Average(R(\textbf{p}_{k_0}^M,k)$ 
    \textbf{for} $\textbf{p}_{k_0}^M \textbf{in} \mathcal{N}_k(\textbf{p}_k^S))$\;
}
$\mathcal{M}_{k+1} \leftarrow \mathcal{M}_{k} \cup \mathcal{S}_k$ \;
\For{$\textbf{p}_{k_0}^M \emph{\textbf{in}} \mathcal{M}_{k+1}$}{
    $R(\textbf{p}_{k_0}^M,k+1) \leftarrow R(\textbf{p}_{k_0}^M,k) + 1$\;
    $P(\textbf{p}_{k_0}^M,k+1) \leftarrow H(\textbf{p}_{k_0}^M,k+1)/R(\textbf{p}_{k_0}^M,k+1)$\;
    \uIf{$P(\textbf{p}_{k_0}^M,k+1)<\theta_p\  \textbf{and}\  R(\textbf{p}_{k_0}^M,k+1)>R_{min}\  \textbf{and}\  H(\textbf{p}_{k_0}^M,k+1)<H_{max}$}{
        Delete($\textbf{p}_{k_0}^M$)\;  
    }
}
\end{algorithm}

\section{Experimental and Evaluation}

We firstly evaluate the performance of \core on the Oxford Radar RobotCar Dataset\cite{RadarRobotCarDatasetICRA2020} which has urban sceneries and variety of weather conditions. Then we evaluate \core on the MulRan\cite{9197298} Dataset and the Boreas Dataset\cite{burnett_boreas22} which have different Radar model, non-urban sceneries and challenging weather conditions. Then, we construct an ablation study to examine the effects of geometry-probability two-step feature filter. To this end, we also implemented the scan-to-frames methods to prove that our map-based method has less drift. All experiments are conducted on a laptop with an AMD Ryzen 9 4800H CPU and 16GB memory. 

\begin{table*}[t]
\caption{Evaluation on Oxford Radar RobotCar dataset}
\center
\begin{tabular}{c|cccccccc|c}

\label{table3}

 Method  & 10-12-32 & 16-13-09 & 17-13-26 & 18-14-14 & 18-15-20 & 10-11-46 & 16-11-53 & 18-14-46 & Mean  \\ \hline
SuMa\cite{8967704}(Lidar)  & 1.1/0.3* & 1.2/0.4* & 1.1/0.3* & 0.9/0.1* & 1.0/0.2* & 1.1/0.3* & 0.9/0.3* & 1.0/0.1* & 1.16/0.3*\\ \hline
ORB\_SLAM2(Vision) & 6.09/1.6 & 6.23/1.7 & 6.41/1.7 & 7.05/1.8 & 11.5/3.3 & 6.11/1.7 & 6.16/1.7 & 7.17/1.9 & 7.09/3.1 \\ \hline
Cen\cite{8793990}  & N/A & N/A & N/A & N/A & N/A & N/A & N/A & N/A & 3.72/0.95\\
Under the Radar\cite{9196835} &  N/A & N/A & N/A & N/A & N/A & N/A & N/A & N/A & 2.05/0.67**\\
Mask by Moving\cite{maskbymoving} &   N/A & N/A & N/A & N/A & N/A & N/A & N/A & N/A & 2.78/0.85\\
HERO\cite{HERO} &   1.77/0.62 & 1.75/0.59 & 2.04/0.73 & 1.83/0.61 & 2.20/0.77 & 2.14/0.71 & 2.01/0.61 & 1.97/0.65 & 1.96/0.66\\
Hong odometry\cite{Hong2021RadarSA} &  2.98/0.8 & 3.12/0.9 & 2.92/0.8 & 3.18/0.9 & 2.85/0.9 & 3.26/0.9 & 3.28/0.9 & 3.33/1 & 3.11/0.9\\
Hong SLAM\cite{Hong2021RadarSA} &  2.17/0.67 & 1.84/0.59 & 2.46/0.81 & 2.21/0.71 & 2.45/0.78 & 2.27/0.9 & 2.24/0.6 & 2.34/0.7 & 2.21/0.7 \\
Hong odometry V2\cite{RadarSLAMv2} &  2.32/0.7 & 2.62/0.7 & 2.27/0.6 & 2.29/0.7 & 2.25/0.7 & 2.16/0.6 & 2.49/0.7 & 2.12/0.6 & 2.32/0.7 \\
Hong SLAM V2\cite{RadarSLAMv2} & 1.98/0.6 & 1.48/0.5 & 1.71/0.5  & 2.22/0.7 & 1.77/0.6 & 1.96/0.7 & 1.81/0.6 & 1.68/0.5 & 1.83/0.6 \\
CFEAR\cite{adolfsson2021cfear} &  1.64/0.48 & 1.86/\textbf{0.52} & 1.66/0.48 & 1.71/0.49 & 1.75/0.51 & 1.65/\textbf{0.48} & 1.99/\textbf{0.53} & 1.79/\textbf{0.50} & 1.76/\textbf{0.50}\\
\core(w/o loop) &   1.80/0.49 & \textbf{1.83}/0.57 & 1.92/0.49 &  1.61/0.49 & 1.69/0.51 & 1.73/0.52 & 1.81/0.58 & 1.92/0.53 & 1.79/0.52\\
\core   &  \textbf{1.63/0.46} & \textbf{1.83}/0.56 & \textbf{1.49/0.47} & \textbf{1.54/0.47} & \textbf{1.61/0.50} & \textbf{1.55}/0.53 & \textbf{1.78}/0.54 & \textbf{1.55/0.50} & \textbf{1.62/0.50}\\\hline
\end{tabular}
\begin{tablenotes}
    \footnotesize
    \item TBALE I shows the evaluation results of different algorithms on Oxford Radar RobotCar Dataset\cite{RadarRobotCarDatasetICRA2020}. \textbf{Bold} font indicate the best results. Mark N/A means the algorithm cannot finish the full sequence. Results marked * cannot be compared directly as these can't run the whole trajectory and fail at some time in the middle. Results marked ** cannot be compared directly as these are trained and evaluated on the same spatial location.
\end{tablenotes}
\end{table*}
\subsection{Performance on Oxford Radar RobotCar Dataset}\label{EOxford}

The Oxford Radar RobotCar Dataset\cite{RadarRobotCarDatasetICRA2020} is an open large-scale radar dataset used for evaluating Radar-based SLAM. It includes 32 sequences of radar data collected while traversing a same route in Oxford with ground truth. The radar data is captured by a Navtech CTS-350X, a Frequency Modulated Continuous Wave (FMCW) scanning radar, which is configured to return 3768 power readings at a resolution of 4.32cm across 400 azimuths and operates at the frequency of 4Hz. For comparison, we selected the same 8 sequences as CFEAR Radarodometry\cite{adolfsson2021cfear} and RadarSLAM\cite{Hong2021RadarSA}. For all sequences, we set $\theta_m=0.25$, $R_{min}=10$ and $H_{max}=10$. The odometry accuracy was measured as proposed in the KITTI odometry benchmark\cite{Geiger2012CVPR} to compute the average translation error (\%) and rotation error (deg/m) over all sub sequences between {100, 200, . . . , 800} m. 

We compared \core with 7 state-of-the-art radar based odometry and SLAM algorithms. 
The evaluation result is shown in TABLE \ref{table3}. 
Experiment shows that \core achieves the best accuracy with a translation of 1.62\% and rotation error of 0.50 deg/100m. The accuracy of \core (1.62\%) is better than CFEAR\cite{adolfsson2021cfear} (1.76\%), RadarSLAM\cite{RadarSLAMv2, Hong2021RadarSA} (1.83\%, 2.21\%), HERO\cite{HERO} (1.96\%), Mask by Moving\cite{maskbymoving} (without CSV, 2.78\%), Under the Radar\cite{9196835} (2.05\%). Without loop closure, the mean accuracy of our odometry (1.79\%) is close to that of the SOTA algorithm CFEAR (1.76\%). When using loop closure, \core imporve by 7.95\% relative to CFEAR.

\subsection{Performance on MulRan Dataset and Boreas Dataset}\label{EMulran}

\begin{table}[t]
\caption{ATE trans. (RMSE) [m] on MulRan Dataset}
\label{table1}
\center
\begin{tabular}{c|c|ccc}
Method & Evaluation & DCC02 & Riverside02 & Mean \\ \hline
PhaRaO\cite{9197231} & & 24.962 & 31.83 & 28.40 \\
SuMa\cite{8967704} & \cite{RadarSLAMv2} & 17.834 & N/A & N/A \\
Hong SLAM \cite{Hong2021RadarSA}& & 24.962 & 95.247 & 60.10 \\
Hong SLAM v2 \cite{RadarSLAMv2}& & 9.878 & 7.049 & 8.46 \\
\core(w/o loop) & & 7.52 & 16.33 & 11.93 \\
\core & & \textbf{5.81} & \textbf{4.85} & \textbf{5.33} \\ \hline

\end{tabular}
\end{table}
\begin{table}[t]
\caption{EVALUATION ON BOREAS DATASET}
\center
\begin{tabular}{c|c|c|c|c}
\label{table5}
 Algorithm   & Sequence & Weather & Error & Mean Error \\ \hline
\multirow{3}{*}{HERO\cite{HERO}} & 01-26-10-59 & snow & 2.00/0.56 & \multirow{3}{*}{2.02/0.56} \\
& 01-26-11-22 & snow & 1.98/0.53 &\\
& 02-09-12-55 & sun  & 2.07/0.59 &\\ \hline
\multirow{3}{*}{\makecell{\core\\(w/o loop)}} & 11-26-13-58 & snow & 1.78/0.44 & \multirow{3}{*}{1.84/0.45} \\
& 12-01-13-26 & snow & 2.02/0.45 &\\
& 12-18-13-44 & sun  & 1.72/0.45 &\\ \hline

\end{tabular}
\end{table}
\begin{figure}[!htp]
	\centering
	\includegraphics[width = \linewidth]{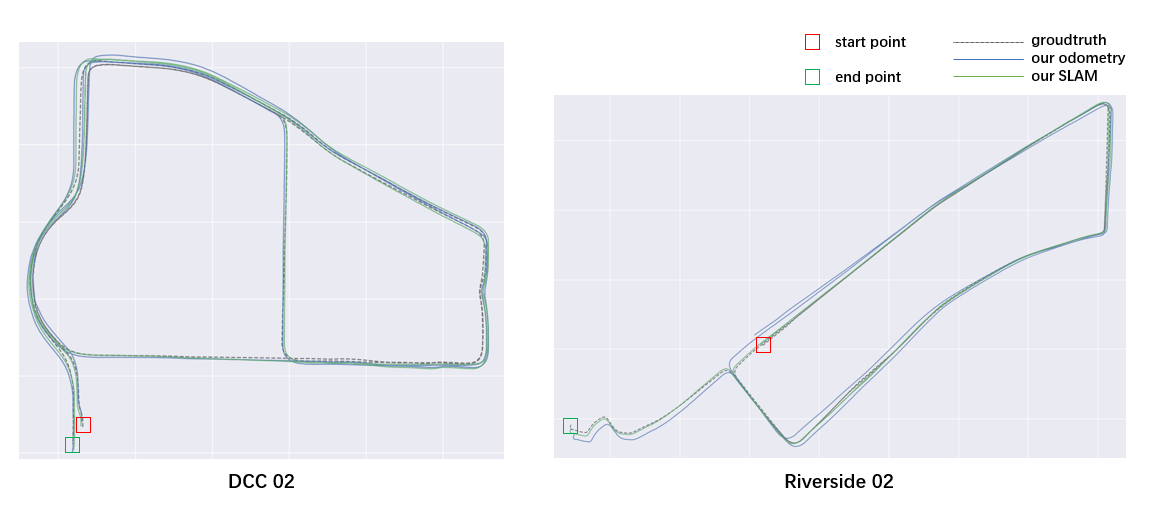}
	\caption{Estimated Odometry Trajectories, SLAM Trajectories and Ground Truth of 2 Sequences from MulRan Dataset\cite{9197298}.}
	\label{fig:mulran}
\end{figure}

\begin{figure}[!htp]
	\centering
	\subfigure[11-26-13-58] {\includegraphics[width = 0.3\linewidth]{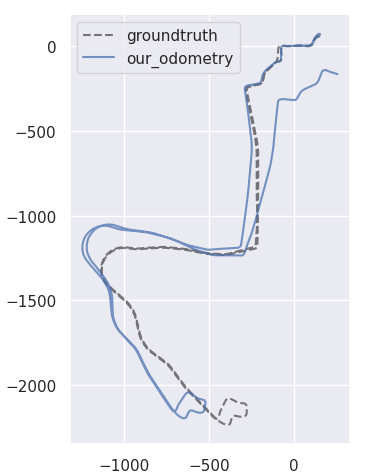}}
	\subfigure[12-01-13-26] {\includegraphics[width = 0.3\linewidth]{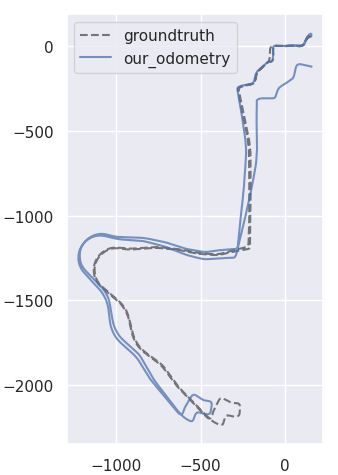}}
	\subfigure[12-18-13-44] {\includegraphics[width = 0.3\linewidth]{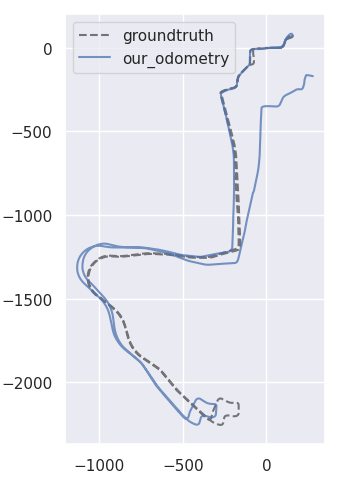}}
	\caption{Estimated odometry (without loop closure) trajectories and groundtruth of 3 sequences from Boreas dataset\cite{9197298}.}
	\label{fig:boreas}
\end{figure}

To test the performance of \core in non-urban scenery and challenging weathers, we evaluate it on MulRan Dataset\cite{9197298} and Boreas Dataset\cite{burnett_boreas22}. The radar images in both datasets are collected from Navtech CIR204-H FMCW scanning radar which is different from Oxford dataset used. This Navtech CIR204-H radar has a range resolution of 0.0596 m, and a total range of 200 m, which is different from Navtech CTS-350X. For both datasets, we set $\theta_m=0.2$, $R_{min}=10$ and $H_{max}=10$. 

\subsubsection{MulRan}the MulRan dataset\cite{9197298} dataset includes four different scenarios of urbans and non-suburbs. In order to compare with PhaRao\cite{9197231} and Hong SLAM\cite{Hong2021RadarSA, RadarSLAMv2}, we selected DCC02 and Riverside02 sequences for evaluation. The third sequence in PhaRao is not open sourced in MulRan. We use the open source tool evo\cite{grupp2017evo} to evaluate the absolute trajectory error (ATE) of the trajectory. 

The trajectory and quantitative results on the sequences are shown in Fig. \ref{fig:mulran} and TABLE \ref{table1}. Experiments show that \core achieves the best accuracy with ATE of 5.33 m. For comparing the accuracy of the odometry (w/o loop), \core(w/o loop) (11.93m) is much better than that of PhaPaO (28.4 m). Compared with the complete SLAM pipeline, the drift error of \core (5.33 m) also exceeds that of Hong SLAM (60.10 m) and Hong SLAM v2 (8.46 m). \core reduces drift error by 37.0\% relative to the SOTA algorithm Hong SLAM v2. Among them, in Riverside02 sequence, which is open, empty and lacks structural textures, our odometry and SLAM algorithms can operate stably where SuMa\cite{8967704} can't finish the full sequence in Hong\cite{RadarSLAMv2}'s evaluation.

\subsubsection{Boreas}The Boreas Dataset\cite{burnett_boreas22} is the latest open source dataset with diverse weather conditions for SLAM and object detection research. In order to compare with HREO\cite{HERO}, we selected three sequences for evaluation. Like HERO's selction, two sequences were taken during a snow day, and the other was taken on a sunny day. The odometry accuracy was measured as proposed in the KITTI odometry benchmark\cite{Geiger2012CVPR}. To be fair, we only tested \core with loop closure disabled.

The trajectory and quantitative results on the sequences are shown in Fig. \ref{fig:boreas} and TABLE \ref{table5}. Experiment shows that \core achieves the accuracy with a translation of 1.84\% and rotation error of 0.45 deg/100m, exceeds HREO's 2.02\% and 0.56 deg/100m. \core reduces translation error by 8.9\% and rotation error by 19.6\% relative to HERO.

\subsection{Ablation Study}\label{Ablation}

\begin{figure}[t]
	\centering
	\includegraphics[width = \linewidth]{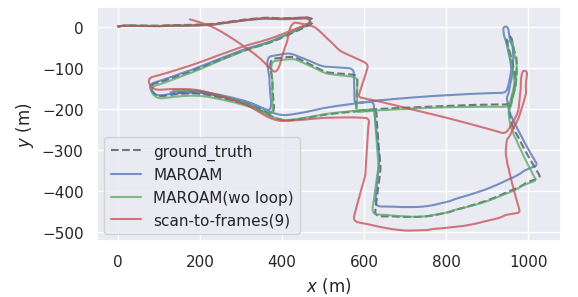}
	\caption{Trajectories of ablation study using sequence "10-12-32" of the Oxford dataset..}
	\label{fig:ablation}
\end{figure}

\begin{table}[t]
\caption{Ablation Experiment}
\center
\begin{tabular}{c|c|c}

\label{table2}

 & Trans.(\%) & Rot.(rad/100m)      \\ \hline

scan-to-frames(5)  & 2.9 & 0.84 \\
scan-to-frames(7)  & 2.57 & 0.74 \\
scan-to-frames(9)  & 2.52 & 0.73 \\
\core(w/o probability filter) & 3.16 & 0.97 \\
\core(w/o geometry filter) & 8.0 & 2.86  \\
\core(w/o loop)  & 1.80 & 0.49\\
\core(w loop)   & 1.63 &  0.46\\ \hline

\end{tabular}

\end{table}

We show how the design of each module in the system affects the performance of the proposed framework using sequence "10-12-32" of the Oxford dataset .The experimental results are shown in TABLE \ref{table2} and Fig. \ref{fig:ablation}.

\subsubsection{Effect of map-based method}We first compare our map-based radar odometry with the scan-to-frames method which using the sum of the feature points from previous frames as the local map. The table and figure item scan-to-frames(n) means the algorithm using $n$ frames as the local map which composed of the surface feature points. Experiments show that the accuracy of our map-based method (1.80\%, 0.49) is 28.6\% higher than that of scan-to-frames (2.52\%, 0.74). 

\subsubsection{Effect of geometry filter}We disable the geometry filter and use the raw feature points to calculate relative pose and build local map and the odometry error increased from 1.80\%, 0.49 rad/100m to 8.00\%, 2.86 rad/100m.

\subsubsection{Effect of probability filter}We disable the probability filter and use the sum of the feature points from all previous frames as the local map. In our experiments, the odometry error increased from 1.80\%, 0.49 rad/100m to 3.16\%, 0.97 rad/100m

\subsubsection{Effect of loop closure}We have done experiments to observe the effect of loop closure on odometry accuracy. Experiment shows that the loop closure has obvious effect on reducing the trajectory drift error. With loop closure, the error dropped from 1.80\%, 0.49 rad/100m to 1.63\%, 0.46 rad/100m.

\section{Conclusion}

In this letter, we propose \core, a framework for map-based radar odometry and mapping for performing state estimation and mapping in complex scenery, weather, and road conditions. The proposed framework uses geometry-probability two-step feature selection in the registration and map update stages. The algorithm first extract surface feature points by calculating the local curvature and local aggregation. A point-to-line ICP method then uses surface feature points to solve the relative pose transformation. Meanwhile, a probability-based filter is used to dynamically filter out the feature points that are not hit frequently in the ICP matching process. We evaluate our \core on three datasets. Experiments show that our algorithm outperforms SOTA algorithm on all three datasets. As devoted contributors to the open-source community, we will open source the algorithm after the paper is accepted.


\bibliographystyle{IEEEtran}
\bibliography{IEEEabrv, references}
\end{document}